\def\arxivversion{}

\documentclass{article} %
\usepackage{iclr2027_conference,times}

\usepackage{amsmath,amsfonts,bm}

\def\eqref#1{equation~\ref{#1}}

\def\1{\bm{1}}

\DeclareMathAlphabet{\mathsfit}{\encodingdefault}{\sfdefault}{m}{sl}
\SetMathAlphabet{\mathsfit}{bold}{\encodingdefault}{\sfdefault}{bx}{n}

\usepackage{hyperref}
\usepackage{url}
\usepackage{booktabs}
\usepackage{graphicx}
\usepackage{siunitx} %
\usepackage{amsmath}

\usepackage{xcolor}

\title{COMPASS: Routing via \textbf{C}apability-\textbf{O}riented\\
\textbf{M}odel \textbf{P}rofiling with \textbf{A}daptive \textbf{S}earch,\\
\textbf{S}election, and Tool Use}

\ifdefined\arxivversion
\author{
\begin{tabular}[t]{c}
\textbf{Amin Rezaei}\thanks{Equal contribution.} \\[2pt]
{\small Invariant Tech Inc.} \\
{\small\texttt{amin@invariant.sh}}
\end{tabular}
\And
\begin{tabular}[t]{c}
\textbf{Pouya Parsa}\footnotemark[1] \\[2pt]
{\small University of Minnesota} \\
{\small\texttt{parsa025@umn.edu}}
\end{tabular}
}

\iclrfinalcopy
\else
\author{Anonymous Authors\\
Anonymous Affiliation\\
\texttt{anonymous@email.com}
}
\fi

\begin{document}

\maketitle
\ifdefined\arxivversion
\lhead{arXiv preprint}
\fi

\begin{abstract}
LLM routing aims to select a capable model at low cost, but specialized routers can require updates as models and tasks change. We propose \textbf{COMPASS}, an agentic router in which an LLM follows a routing recipe, gathers performance and cost evidence through tools, and optionally inspects candidate answers before selecting a model. Our evaluations on LLMRouterBench and DeepSWE show that agentic routing can improve accuracy and lower model-call cost relative to specialized and trained routing baselines, without routing-specific training. On LLMRouterBench, COMPASS reaches 69.85\% accuracy at \$139.57, compared with RLCascadeRouter's 69.14\% at \$180.22. On DeepSWE under fixed five-fold cross-validation, it reaches 78.1\% at \$215, compared with 74.1\% at \$500 for the best single model. COMPASS also supports routing across benchmarks and adapts to an expanded model pool by reading new evaluation records, without changing the router's weights or prompt.

\end{abstract}

\section{Introduction}
\label{sec:intro}

The ecosystem of large language models has fragmented into distinct capability--price tiers. Frontier models dominate benchmarks but are expensive and often slow; small open-weight models are cheap and fast, and for many task distributions they are entirely sufficient. Consider a user requesting a simple Python script: this task does not require a frontier model, and a cost-effective model would complete it at a fraction of the price. The difficulty is that this knowledge is available only \emph{in hindsight}: one would need to query both the cheap and the frontier model to know whether the cheap model actually sufficed. The routing problem is precisely the problem of estimating this counterfactual \emph{before} committing to a model.

Routing must navigate a three-way trade-off:
\begin{itemize}
\item \textbf{Cost.} Cheaper models reduce per-token spend, but only if they succeed; failed attempts followed by escalation can cost more than routing correctly the first time.
\item \textbf{Latency.} A cheap model with low throughput that emits many tokens can take far longer end-to-end than a stronger model that solves the task in substantially fewer tokens. Cost-optimal and latency-optimal choices frequently diverge.
\item \textbf{Accuracy.} Some users simply want the strongest available response regardless of price.
\end{itemize}

Most prior work optimizes a single axis, usually cost reduction at near-constant quality, because cost is the easiest axis to measure and experiment on \citep{chen2024frugalgpt,ong2025routellm}. Optimizing for accuracy instead pushes evaluation toward frontier models, where per-experiment API costs are substantial and public fine-grained evaluation data is scarce.

\paragraph{Routing recipes.} Users can specify what they value without estimating how difficult a query will be. A performance-focused recipe asks the router to prioritize reliable completion; a cost-optimized recipe asks it to seek cheaper configurations when the evidence supports comparable quality. Qualitative levels such as low, medium, or high can express the strength of these preferences. The recipe provides guidelines for gathering evidence and making a decision, while the router assesses the query using that evidence. Numerical trade-off parameters can still be used inside tools and experimental sweeps; users need not tune them for each request.

\paragraph{The staleness problem.} When model-specific performance estimates are encoded in a trained router, changes in the model pool or task distribution can require new supervision or updates. Methods that support unseen models reduce this burden, but still need useful evidence about their capabilities. We instead keep evaluation outcomes in an external corpus that the router can query at inference time. New models can be added by collecting and inserting their evaluation records, without updating the routing agent's weights. This shifts the maintenance burden to evidence collection, whose cost and coverage remain important limitations.

\paragraph{Our proposal.} We separate \emph{routing knowledge} from \emph{routing skill}. Knowledge about which model scored what, at what cost, and on which task lives in an external evaluation corpus. The skill to query that corpus and decide under a routing recipe comes from the tool-use capabilities of an existing LLM. COMPASS requires no routing-specific weight training: the agent follows the recipe, adaptively explores evidence, and submits a result when it judges the evidence sufficient. This yields three design requirements:
\begin{enumerate}
\item \textbf{Recalibration through evidence.} New models and benchmarks are incorporated by inserting evaluation records into the corpus; the router's weights are untouched.
\item \textbf{Recipe-oriented routing.} The router follows input guidelines, including performance-focused and cost-optimized recipes with qualitative preference levels.
\item \textbf{Low routing overhead.} Evidence acquisition and agent inference must be economical enough to preserve any savings in model execution.
\end{enumerate}

\paragraph{Contributions.}
\begin{enumerate}
\item We formulate routing as evidence gathering through tools, separating updatable model records from an existing LLM's decision-making capabilities (Section~\ref{sec:method}).
\item We implement COMPASS, which follows routing recipes to gather evidence, inspect candidate answers, and select a model or stop unresolved (Section~\ref{sec:compass}).
\item We evaluate COMPASS against routing baselines on LLMRouterBench and under five-fold cross-validation on DeepSWE. COMPASS reaches 78.1\% accuracy on DeepSWE and 76.7\% using only TerminalBench evidence. Expanding the model pool improves accuracy and lowers model-call spend without retraining; ablations measure the effects of router choice and model names (Section~\ref{sec:experiments}).
\end{enumerate}

\section{Related Work}
\label{sec:related-work}

\paragraph{Trainable model routing.}
One-step routers learn to select a model or execution configuration from preferences and difficulty estimates~\citep{ong2025routellm,ding2024hybrid}, predicted performance~\citep{xue2026r2,fernandez2026radar,liu2024optllm,zhuang2024embedllm}, query--model representations~\citep{chen2024routerdc,feng2024graphrouter,jitkrittum2026uniroute}, online feedback~\citep{panda2025pilot}, or latent expert interactions~\citep{feinashley2025mixture}. They commit before inspecting a completed answer and therefore cannot revise that query's route using post-generation evidence. Sequential systems can instead purchase more precise estimates~\citep{fisch2026pandora}, inspect an answer before stopping or escalating~\citep{chen2024frugalgpt,shah2025selectthenroute,chuang2024selfref,dekoninck2025cascade,gupta2024lmcascades,aggarwal2024automix}, or learn multi-call orchestration~\citep{nielsen2026conductor,shao2025routeandreason}. Across both families, trained predictors, scorers, and policies may need refreshed outcomes or updates as model pools and task distributions change. Some methods explicitly support new models, but the reliability of transfer across pools remains an empirical question.

\paragraph{Training-free model routing.}
Training-free routers reuse stored query--model outcomes without fitting a dedicated neural policy. The Avengers, Avengers-Pro, and UniRoute's k-means variant summarize model performance in prompt clusters~\citep{zhang2025avengers,zhang2025avengerspro,jitkrittum2026uniroute}. Nearest-neighbor variants in RouterBench and CARROT, together with benchmark-based and online kNN routers, transfer observed success or inference-cost estimates from similar evaluated queries~\citep{hu2024routerbench,shnitzer2023benchmark,somerstep2025carrot,li2025rethinking,wu2025trainingfree}. Eagle blends global and local Elo rankings~\citep{zhao2024eagle}, while ProxRouter reweights nearby evidence to improve outlier robustness~\citep{patel2025proxrouter}. The quality of these estimates depends on the coverage and relevance of stored examples: when neighbors are sparse, noisy, or close in embedding space yet different in difficulty or required capability, local model rankings can be unstable or misleading. Global rankings and proximity weights can reduce this risk, but a weakly supported query may still require other evidence.

These approaches expose two sources of fragility: trainable predictors and policies may require new supervision or retraining as model pools change, and their transfer to unseen pools remains uncertain; neighbor-based training-free estimates can become unreliable when relevant examples are sparse, weakly related, or noisy. COMPASS separates model-specific knowledge from routing skill: outcomes and inference costs reside in an updatable evaluation corpus, while an existing LLM agent uses anonymized model identities to decide which evidence to acquire. Adding a model requires calibration outcomes rather than changes to the agent's weights. Neighbor retrieval is only one optional evidence tool; the agent can also consult model profiles and quality--cost estimates, inspect a generated answer, request further evidence, or stop unresolved. More informative tools can enrich this search as they become available, provided their inference cost and effect on final-answer reliability are assessed.

\section{Method}
\label{sec:method}

\paragraph{Routing Problem.}
Let $(X,Y)\sim P$ denote a task and its success criterion. The finite pool
$\mathcal M$ is fixed across queries within a deployment and throughout each
routing episode, but may differ from the pool used to develop the router.
Each $m\in\mathcal M$ specifies a model and its execution settings, including
prompt, decoding, tools, and reasoning budget. Executing $m$ yields a possibly
random answer $\widehat Y_m$ at realized inference cost $C_m$. With binary
failure indicator $L_m$, define
\begin{equation}
\begin{aligned}
 s(x,m)&:=\Pr(L_m=0\mid X=x),\\
 \kappa(x,m)&:=\mathbb E[C_m\mid X=x].
\end{aligned}
 \label{eq:method-success-cost}
\end{equation}
For a system-defined completion target $\rho\in(0,1]$, the ideal single-call
router selects
\begin{equation}
 r^*_{\rho}(x,\mathcal M)\in
 \underset{m\in\mathcal M:\,s(x,m)\geq\rho}{\operatorname{arg\,min}}
 \kappa(x,m).
 \label{eq:method-single-call}
\end{equation}
This is a completion-first objective: among configurations meeting the target,
it chooses the cheapest in expected inference cost, without a user-specified
spending cap. If no configuration qualifies, a single preselected execution
cannot meet the target. The true $s(x,m)$ and $\kappa(x,m)$ are unknown to the
router.

\paragraph{Sequential Evidence Acquisition.}
The agentic router can inspect evidence and candidate answers before returning
one. After $t$ interactions, its history is
$h_t=(x,\mathcal M,(a_1,o_1),\ldots,(a_t,o_t))$, where $a_i$ is an action and
$o_i$ its observation. If $K_t$ answers have been previewed, the router policy
$\pi_r$ chooses from
\begin{equation}
\begin{aligned}
\mathcal A(h_t)={}&\mathcal A_{\rm ev}(h_t)
 \cup\{\operatorname{preview}(m):m\in\mathcal M\}\\
 &\cup\{\operatorname{accept}(j):1\leq j\leq K_t\}
 \cup\{\operatorname{unresolved}\}.
\end{aligned}
\label{eq:method-actions}
\end{equation}
Evidence actions may retrieve outcomes from similar tasks, query model
profiles or predictive routers, and assess a previewed answer. A preview runs
configuration $m$ and stores its answer for possible acceptance; accepting it
does not run $m$ again. Tool availability and call limits restrict
$\mathcal A(h_t)$.

\paragraph{Objective and Stopping.}
Let $\tau$ be the terminal interaction, and let $c_i$ be its incremental
inference cost, including router reasoning, evidence calls, and previews.
Let $Z_r$ indicate whether the returned answer completes the task, with
$Z_r=0$ for an unresolved outcome. We seek a finite-terminating policy that
solves
\begin{equation}
\begin{aligned}
 \min_{\pi_r}\quad&\mathbb E_{\pi_r}\!\left[\sum_{i=1}^{\tau}c_i\right]\\
 \text{s.t.}\quad&\Pr_{\pi_r}(Z_r=1\mid X=x)\geq\rho.
\end{aligned}
\label{eq:method-cascade}
\end{equation}
The constraint applies for $P_X$-almost every $x$ and concerns the final
answer, not the marginal success rate of the model that generated it. The cost
counts all calls, including evidence queries and rejected previews.

At history $h_t$, the ideal remaining-cost comparison for an action $a$ is
\begin{equation}
 W(a\mid h_t):=
 \inf_{\pi\in\Pi_\rho(h_t,a)}
 \mathbb E^{\pi}\!\left[
   \sum_{i=t+1}^{\tau}c_i\,\middle|\,h_t
 \right],
 \label{eq:method-continuation}
\end{equation}
where $\Pi_\rho(h_t,a)$ contains admissible policies starting with $a$ that
meet the completion target conditional on $h_t$; $W=+\infty$ if none exists.
Past costs are sunk in this comparison. The router need not compute $W$ or
know the true success probabilities: it uses the available evidence to decide
whether to accept an answer, gather more evidence, preview another model, or
stop unresolved. Its confidence estimates may be implicit or uncalibrated, so
the completion target must be evaluated using actual final outcomes.

\section{COMPASS}
\label{sec:compass}

\begin{figure}[t]
\centering
\includegraphics[width=\linewidth]{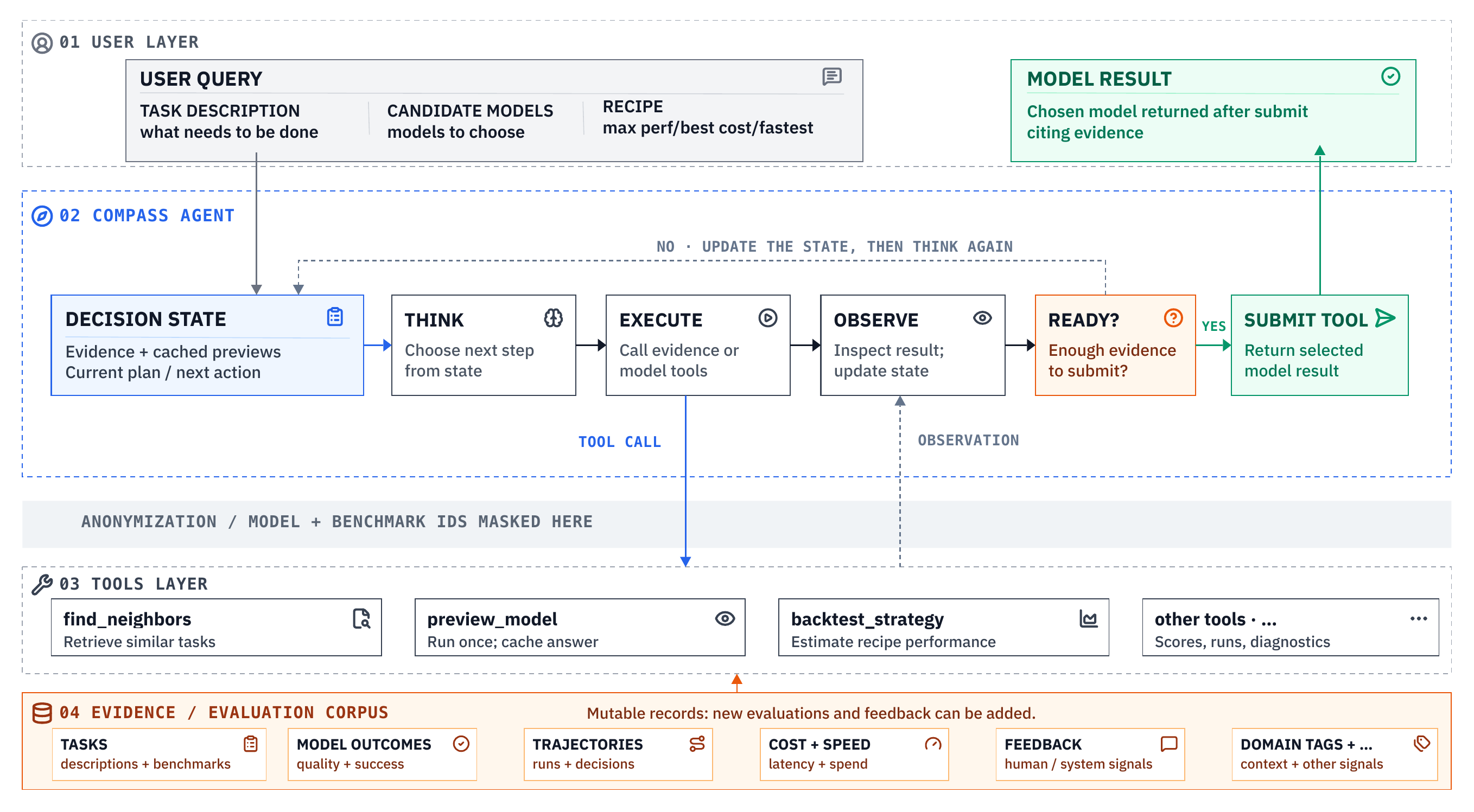}
\caption{COMPASS queries evaluation records, optionally pays to preview an answer, and selects a model and reasoning-effort level. New model records can be added without retraining the router.}
\label{fig:arch}
\end{figure}

COMPASS gives an existing LLM a routing prompt (a \emph{recipe}), evaluation corpus $\mathcal{E}$, candidate list, and accuracy--cost preference $\alpha\in[0,1]$. For each query, it gathers evidence and submits a configuration $(m,e)$ (a model alias and reasoning-effort level) with free-text evidence and a self-reported, uncalibrated confidence label. Different effort levels are separate choices with measured pass rates and costs. No routing-specific weights are trained.

\subsection{Evidence and tools}
\label{sec:corpus}

A per-run access mask fixes the visible tasks, candidates, and aliases. The corpus stores each attempt's outcome, USD cost, token usage, and available output (patch, transcript, or answer). DeepSWE has roughly four attempts per cell; LLMRouterBench has one. A preview therefore exposes one sample on DeepSWE but the sole scored answer on LLMRouterBench. Task descriptions are embedded once for nearest-neighbour retrieval, restricted to visible tasks.

For each mask, we precompute configuration pass rates and costs, ecosystem-specific rates, the accuracy--cost frontier, and recommendations over the $k$ nearest tasks. Diagnostics estimate task-specific routing gains and how well an output judge predicts scores. Policy comparisons hold out each visible task in turn and report paired confidence intervals against the cheapest policy. Table~\ref{tab:tools} lists the interface.

\begin{table}[t]
\caption{COMPASS tools. Evidence reads are free; answer previews add model-call cost.}
\label{tab:tools}
\centering
\small
\begin{tabular}{llp{1.75in}}
\toprule
\bf Tool & \bf Input/output & \bf Purpose \\
\midrule
\texttt{find\_similar\_tasks} & $q \mapsto \{t_i\}$ & Nearest visible tasks \\
\texttt{get\_task\_description} & $t \mapsto \text{text}$ & Task text shown to models \\
\texttt{lookup\_scores} & $(\{t\},\{m\}) \mapsto \{(s,c)\}$ & Outcomes and costs; optional grouping \\
\texttt{list\_configurations} & $\emptyset \mapsto \{(m,e)\}$ & Allowed candidates \\
\texttt{neighbourhood\_evidence} & $q \mapsto$ table & All candidates' neighbour outcomes and benchmark composition \\
\texttt{reliability\_table} & $\emptyset \mapsto$ table & Overall and ecosystem pass rates, costs, frontier \\
\texttt{statistic\_recommendation} & $\emptyset \mapsto (m,e)$ & kNN choice at $\alpha$ \\
\texttt{benchmark\_diagnostics} & $\emptyset \mapsto$ (gain, AUC) & Routing gain over best constant; judge accuracy \\
\texttt{backtest\_policy} & $\{\pi\} \mapsto \{(\bar{s},c,\Delta,\mathrm{CI})\}$ & Held-out policy scores and paired comparisons \\
\midrule
\texttt{preview\_answer} & $(m,e) \mapsto \tau$ & Paid stored attempt on current task \\
\texttt{submit\_recommendation} & $(m,e,\text{evidence},\text{conf.})$ & End session \\
\bottomrule
\end{tabular}
\end{table}

Tools have individual call limits; identical calls are deduplicated, and recipes allow 4--16 calls total. Reading corpus evidence adds no model-call cost. Each preview is charged even if rejected. Precomputed tables reduced calls from six to two or three per task while maintaining or improving accuracy: the agent compares evidence, decides whether to preview, and stops instead of rebuilding statistics.

\subsection{Routing recipes}
\label{sec:recipes}

Recipes specify the objective, tool order, and selection rule. We evaluate four families:
\begin{itemize}\setlength\itemsep{0pt}
\item \textbf{Single pick:} read reliability or neighbourhood evidence, choose from its accuracy--cost frontier using $\alpha$, and submit; one or two calls, no previews.
\item \textbf{Cascade:} start with the kNN recommendation, inspect its answer, and accept or try the next candidate (Section~\ref{sec:method}).
\item \textbf{Conditional preview:} allow all evidence sources, but enable previews only under preset conditions, such as tied leaders, conflicting evidence, or benchmark diagnostics.
\item \textbf{Choose a policy:} compare predefined policies with \texttt{backtest\_policy}, then apply one. The agent cannot invent policies.
\end{itemize}
Shared prompt segments keep selection rules identical across arms. Paraphrasing the single-pick rule changed DeepSWE pass@1 by 5.5 points; replacing its informal trade-off with an executable objective changed it by 9.5 points.

\subsection{Model identities and updates}
\label{sec:antimem}

Aliases are generated per run and resolved outside the agent. They hide names and reputation, although scores, prices, or output style may reveal identity (Section~\ref{sec:abl-anonymity}). Adding candidates requires only their records and a candidate-list update, with the router's weights unchanged (Section~\ref{sec:exp-generalisation}).

\subsection{Execution and evaluation}

Sessions end on submission or budget exhaustion; exhausted sessions are unresolved and retried only under the transient-failure policy. Task cost includes every previewed configuration and the submitted configuration at their measured per-attempt costs; router inference is reported separately. Corpus-only recipes score the chosen configuration's mean pass rate; cascades score the inspected, accepted attempt. We report these scoring conventions separately (Section~\ref{sec:experiments}).

Evidence and evaluation tasks are disjoint. Visible records supply COMPASS's evidence and baseline training data, never router weight updates. Tables are rebuilt per fold; removing held-out tasks must leave them byte-identical. We study one routing decision per task, not model switching within a conversation.

\section{Experiments}
\label{sec:experiments}

\subsection{Evaluation data}
\label{sec:exp-data}

Routing cannot be evaluated from leaderboard aggregates. A router makes a decision per task, so evaluating it requires, for every task and candidate model, whether the model solved the task and how much the attempt cost. This per-task record is the scarce resource and determines which benchmarks we can use.

Our main routing evaluations use three sources. \textbf{LLMRouterBench}~\citep{li2026llmrouterbench} provides pre-collected per-prompt, per-model results for thirteen models across ten benchmarks: AIME~\citep{maaAIME}, ArenaHard~\citep{li2024arenahard}, GPQA~\citep{rein2023gpqa}, Humanity's Last Exam (HLE)~\citep{phan2025hle}, LiveCodeBench~\citep{jain2024livecodebench}, LiveMathBench~\citep{liu2024livemathbench}, MMLU-Pro~\citep{wang2024mmlupro}, SimpleQA~\citep{wei2024simpleqa}, SWE-bench~\citep{jimenez2023swebench}, and $\tau^2$-Bench (TAU2)~\citep{barres2025tau2bench}. We use its released record dump with the performance--cost configuration's filters. It is also the corpus used for our head-to-head baseline comparison. Its main limitation is that it contains \emph{one record per prompt and model}: there are no repeated trials, so pass@$k$ and per-task variance cannot be measured.

\textbf{DeepSWE}~\citep{huang2026deepswe} and \textbf{TerminalBench}~\citep{merrill2026terminalbench} publish traces. Their repeated trials per cell (four on DeepSWE) make the pass@$k$ and ceiling analysis in \S\ref{sec:exp-ceiling} possible, since we can split trials into disjoint evidence and evaluation sets. The traces also let the router inspect what a model did rather than only whether it passed. Our DeepSWE catalog contains \num{31617} trials over 113 tasks and 28 model/effort configurations. The paired TerminalBench cohort contains 66 tasks on an exact 18-configuration roster.

For the ceiling analysis in Section~\ref{sec:exp-ceiling}, we also use repeated-trial records from \textbf{FrontierSWE}~\citep{kondra2026frontierswe}.

\subsection{LLMRouterBench: head-to-head on a 70/30 split}
\label{sec:exp-llmrouterbench}

We reproduce the LLMRouterBench split with seed 42, using 70\% of tasks for training and 30\% for testing. The three lowest-cost COMPASS points in Figure~\ref{fig:llmrouterbench} use pilot measurements with a $+0.5$ percentage-point display adjustment and costs scaled to the test-set size.

\subsubsection{Baselines}
\label{sec:exp-baselines}

We run FrugalGPT, GraphRouter, and Avengers-Pro on this split using stored records and upstream implementations where available. We report our measured results for these baselines and include one published RLCascadeRouter operating point for comparison. FrugalGPT uses thirteen per-model DistilBERT answer scorers and a shared scorer, with cascade orders and thresholds searched over \num{1885} ordered cascades. GraphRouter uses the released GNN configuration with seed 1881 and \num{2000} epochs. Avengers-Pro uses $k$-means over the \num{8558} training embeddings and top-$k$ cluster routing.

\subsubsection{Results}

\begin{figure}[t]
\centering
\includegraphics[width=\linewidth]{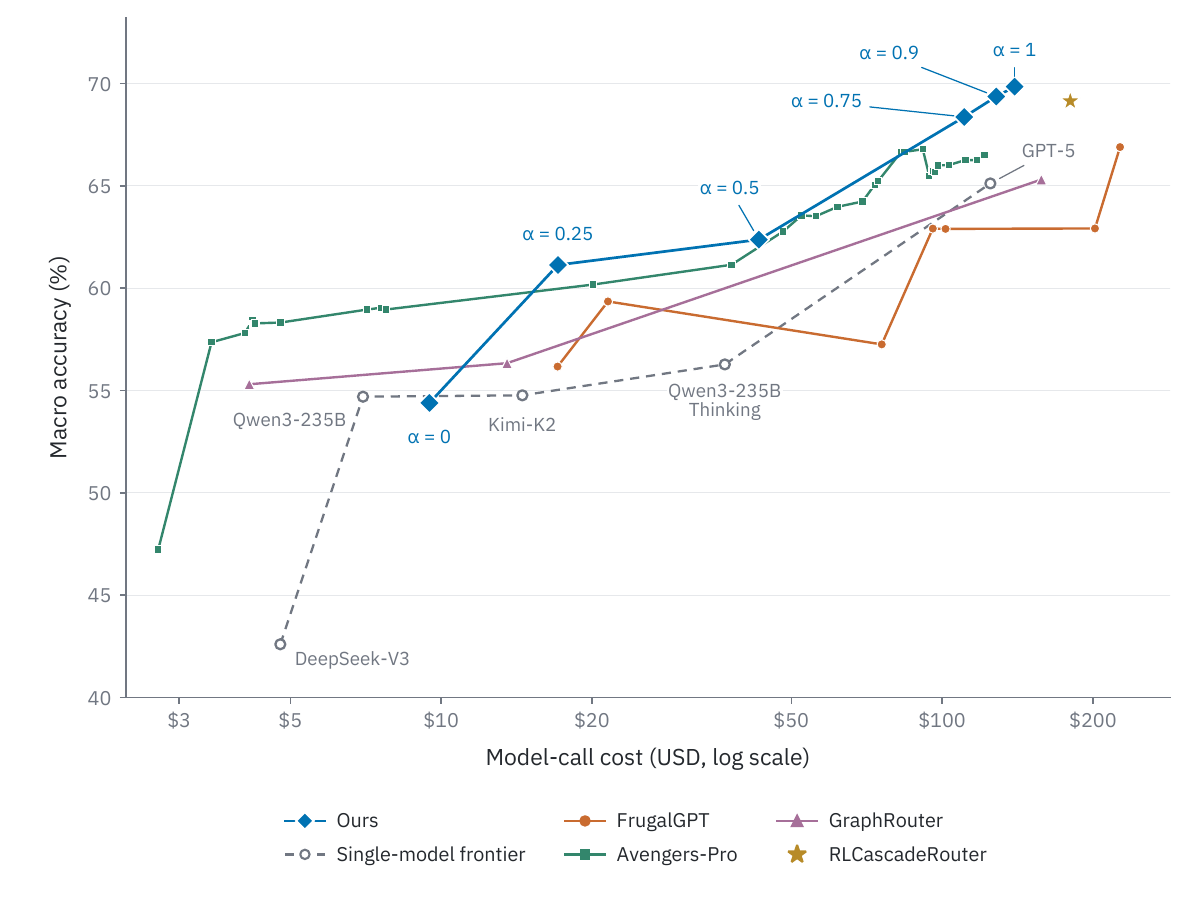}
\caption{Accuracy and model-call cost of COMPASS, single models, and routing baselines on LLMRouterBench.}
\label{fig:llmrouterbench}
\end{figure}

Figure~\ref{fig:llmrouterbench} shows how COMPASS trades model-call cost for accuracy. Its most accurate setting reaches \textbf{69.85\% at \$139.57}, compared with the plotted RLCascadeRouter reference at 69.14\% and \$180.22. COMPASS also reaches 69.36\% at \$128.30, exceeding GPT-5's 65.12\% at a similar cost of \$124.82.

At a lower cost, COMPASS reaches 68.37\% at \$110.69. The best recorded Avengers-Pro setting reaches 66.79\% at \$91.50, while FrugalGPT reaches 66.89\% at \$226.54 and GraphRouter reaches 65.31\% at \$157.79. COMPASS therefore provides the highest accuracy in the figure, while Avengers-Pro remains competitive at lower budgets. These costs exclude router inference.

\subsection{DeepSWE: five-fold cross-validation}
\label{sec:exp-cv}

DeepSWE has only 113 tasks, so a 70/30 split leaves a small test set and produces highly variable results. We therefore use five-fold cross-validation, holding each evaluation task out of the routing evidence and keeping COMPASS's policy fixed across folds. Accuracy averages the recorded trial outcomes: one success in four trials contributes 25\%.

\begin{figure}[t]
\centering
\begin{minipage}[t]{0.49\linewidth}
\centering
\includegraphics[width=\linewidth]{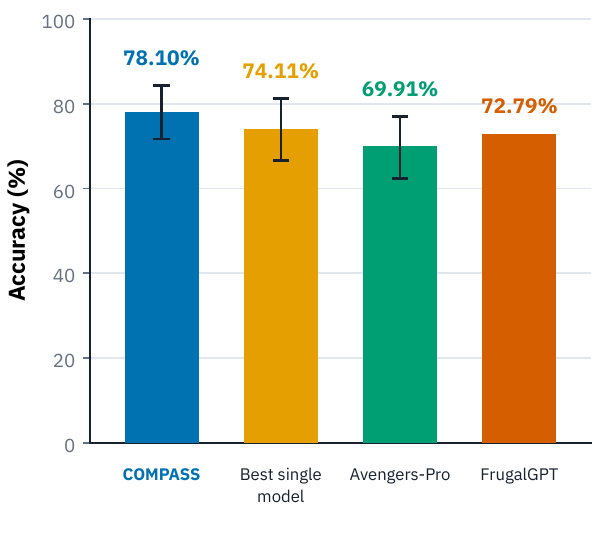}
\small (a) DeepSWE
\end{minipage}\hfill
\begin{minipage}[t]{0.49\linewidth}
\centering
\includegraphics[width=\linewidth]{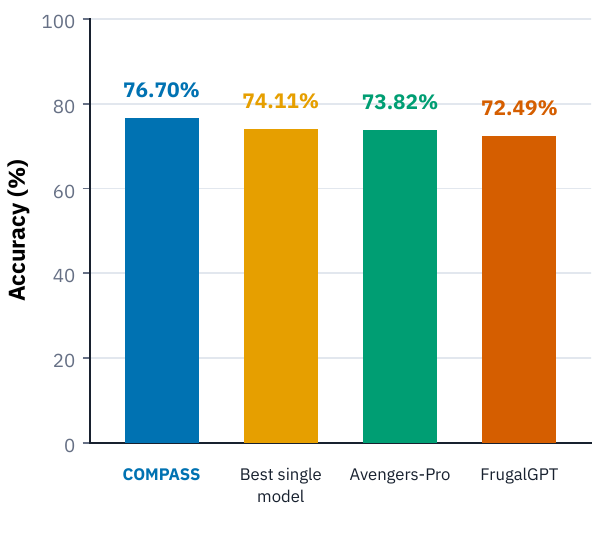}
\small (b) TerminalBench $\rightarrow$ DeepSWE
\end{minipage}
\caption{DeepSWE accuracy with in-domain evidence (left) and TerminalBench evidence (right). Error bars show available 95\% task-bootstrap intervals.}
\label{fig:cv-frontier}
\label{fig:ood-tb-ds}
\end{figure}

Figure~\ref{fig:cv-frontier}a shows COMPASS at \textbf{78.1\%}, compared with 74.1\% for the best single model (\texttt{gpt-6-astra/xhigh}), 69.9\% for Avengers-Pro, and 72.8\% for FrugalGPT. COMPASS uses \$215 in model-call spend, compared with \$500 for the best single model. The accuracy estimates favor COMPASS, although the confidence intervals remain wide on this small task set.

\subsection{Out-of-distribution routing}
\label{sec:exp-ood}

Here the router receives evidence from \emph{one benchmark only} and must route tasks from a different benchmark. We use TerminalBench evidence to route 113 DeepSWE tasks, with three routing repeats and the same 18 model/effort configurations in both cohorts.

With TerminalBench evidence, COMPASS reaches 76.7\% accuracy on DeepSWE, compared with 74.1\% for the best single model, 73.8\% for Avengers-Pro, and 72.5\% for FrugalGPT (Figure~\ref{fig:ood-tb-ds}b). The paired confidence interval for COMPASS's advantage over the best single model includes zero. The result shows that evidence from another benchmark can support useful routing, but does not establish an accuracy improvement over that model.

\subsection{How much can routing improve accuracy?}
\label{sec:exp-ceiling}

Routing helps only when different models are reliably better on different tasks. To estimate this opportunity, we use repeated trials on DeepSWE and FrontierSWE: one group of trials selects a model, and a disjoint group evaluates it. Across 60 random splits, we compare selection using evidence from the target task itself, the most behaviorally similar other task, and a random task against always using the best single model.

On DeepSWE, evidence from the task itself yields a 9.0-point gain, while evidence from the most similar other task yields 4.0 points (Figure~\ref{fig:ceiling} in Appendix~\ref{app:routing-ceiling}). This suggests useful routing opportunities, but only part of that opportunity is accessible through related tasks. On FrontierSWE, neither approach beats the best single model, leaving little measured benefit from per-task selection.

The point is to distinguish apparent oracle gains from gains a router can actually recover. Choosing and scoring a model on the same outcomes exaggerates the opportunity by selecting lucky trials. LLMRouterBench has only one record per task and model, so it cannot support this disjoint-trial check; its oracle gap alone is not evidence of achievable routing gains.

\subsection{Generalization to a new model pool}
\label{sec:exp-generalisation}

We expand the DeepSWE catalog from 18 to 28 models by adding their evaluation records, then rerun COMPASS on the same 113 tasks under fixed five-fold cross-validation. With \texttt{gpt-5.6-luna/medium} as the router, accuracy rises from 63.9\% to \textbf{78.1\%}, while model-call spend falls from \$474 to \textbf{\$215}. The routing model, prompt, and policy remain fixed: COMPASS adopts the new models directly from their records, without retraining.

\subsection{Ablations}
\label{sec:ablations}

\subsubsection{Does anonymity matter?}
\label{sec:abl-anonymity}

We compare opaque aliases with real model names while keeping the router, prompt, candidate pool, and tasks fixed. On LLMRouterBench, revealing names changes macro accuracy by $-0.49$ points for \texttt{gpt-5.6-luna} and $+0.47$ points for DeepSeek-V4.1-Flash. These single-run comparisons show small, opposite changes, with no consistent accuracy benefit from revealing model identities.

Names nevertheless affect the choices: Luna changes 19\% of its selections, compared with 6\% for DeepSeek. Both shift slightly toward GPT-5, whose selection share rises from 35.7\% to 37.8\% for Luna and from 33.6\% to 34.1\% for DeepSeek. This suggests that model-specific priors can influence routing even when aggregate accuracy barely changes. Anonymization removes explicit model names as a cue for memorized reputation or flagship preference, encouraging decisions based on the supplied evidence.

\subsubsection{How much does the routing model matter?}
\label{sec:abl-router-model}

We swap the routing LLM while keeping the agent, prompt, objective, candidate pool, and tasks fixed. On the LLMRouterBench ablation, \texttt{gpt-5.6-luna}, \texttt{gpt-5.6-sol}, and DeepSeek-V4.1-Flash differ by only 1.15 macro accuracy points. DeepSeek selects models with 22\% less model-call spend than Luna, at a cost of 0.86 accuracy points. Changing the router therefore affects spending more than accuracy in this setting; these costs exclude the router's own inference bill.

Because these LLMs have not been trained for our routing objective, they can interpret the same cost--accuracy preference differently. DeepSeek tends to favor cheaper configurations, which can sacrifice accuracy when the model pool offers a wider range of choices. On DeepSWE with fixed five-fold CV, it reaches 68.4\% at \$105, compared with Luna's 78.1\% at \$215. Router choice thus matters more in this setting: lower model-call spend can come with a substantial accuracy loss.

\section{Limitations}
\label{sec:limitations}

\begin{itemize}
\item \textbf{Existing LLMs as routing agents.} COMPASS uses an existing LLM; we have not trained a specialized model for routing. The results motivate investigating such training as a next step, particularly to improve evidence use and reduce inference overhead, but do not establish that it will improve performance.
\item \textbf{Evaluation cost and coverage.} Collecting per-task outcomes and trajectories across models is expensive. Routing quality depends on the coverage, reliability, and relevance of those records. Adding a model requires new evidence even though it does not require retraining the agent, and sparse or mismatched data can lead to poor choices.
\item \textbf{Distribution-dependent routing.} Different task distributions have different routing profiles, and the same router does not perform best across all datasets. Strong results on one cohort do not establish reliable transfer to another.
\item \textbf{Routing overhead and uncertainty.} Agent inference can consume the savings from cheaper model choices. DeepSWE estimates have wide confidence intervals, and most experiments use one routing realization per task. The reported task-bootstrap intervals therefore do not capture all routing stochasticity.
\end{itemize}

\section{Conclusion}
\label{sec:conclusion}

COMPASS uses an existing LLM to follow routing recipes and adaptively gather evidence from an updatable evaluation corpus. This separation lets new models be incorporated without retraining the routing agent. Experiments on LLMRouterBench and DeepSWE demonstrate useful accuracy--cost tradeoffs, while the routing-cost analysis and cross-benchmark transfer results show that gains depend on both the evidence and the task distribution.

A next step is cache-aware multi-turn routing, in which the router revisits its decision as a session reveals more about the task. Switching models can incur fresh context-processing costs and lose caching benefits, so a useful policy must weigh these costs against the expected benefit of continuing, switching, or delegating a subtask. Evaluating such policies requires session trajectories with recorded switching decisions, outcomes, and costs. Constructing and evaluating this data is future work; the present experiments address individual tasks.

\subsection*{AI use statement}

We used generative AI tools to aid and polish writing, including drafting,
editing, and \LaTeX{} formatting; for retrieval and discovery of relevant
literature; and for research execution, including software development and
faster experimentation. These activities were carried out under human
oversight. The authors reviewed and revised AI-assisted text, checked cited
work against original sources, and verified AI-assisted code, experimental
outputs, and analyses. We take responsibility for the final content of this
work, including its claims, results, and artifacts.

\subsection*{Reproducibility statement}

The routing objective and action space are specified in Section~\ref{sec:method},
and the COMPASS architecture and tool interface in Section~\ref{sec:compass}.
Section~\ref{sec:experiments} describes the evaluation corpora, the seed-42 split,
baseline implementations, and cross-validation and transfer protocols.
COMPASS uses existing LLMs without routing-specific weight training.
Baseline configurations and the single RLCascadeRouter reference point
are documented in Section~\ref{sec:exp-baselines}; routing overhead and
statistical limitations are discussed alongside the results and in
Section~\ref{sec:limitations}.

\bibliography{iclr2027_conference}
\bibliographystyle{iclr2027_conference}

\clearpage
\appendix
\section{Routing accuracy ceiling}
\label{app:routing-ceiling}

\begin{figure}[htbp]
\centering
\includegraphics[width=\linewidth]{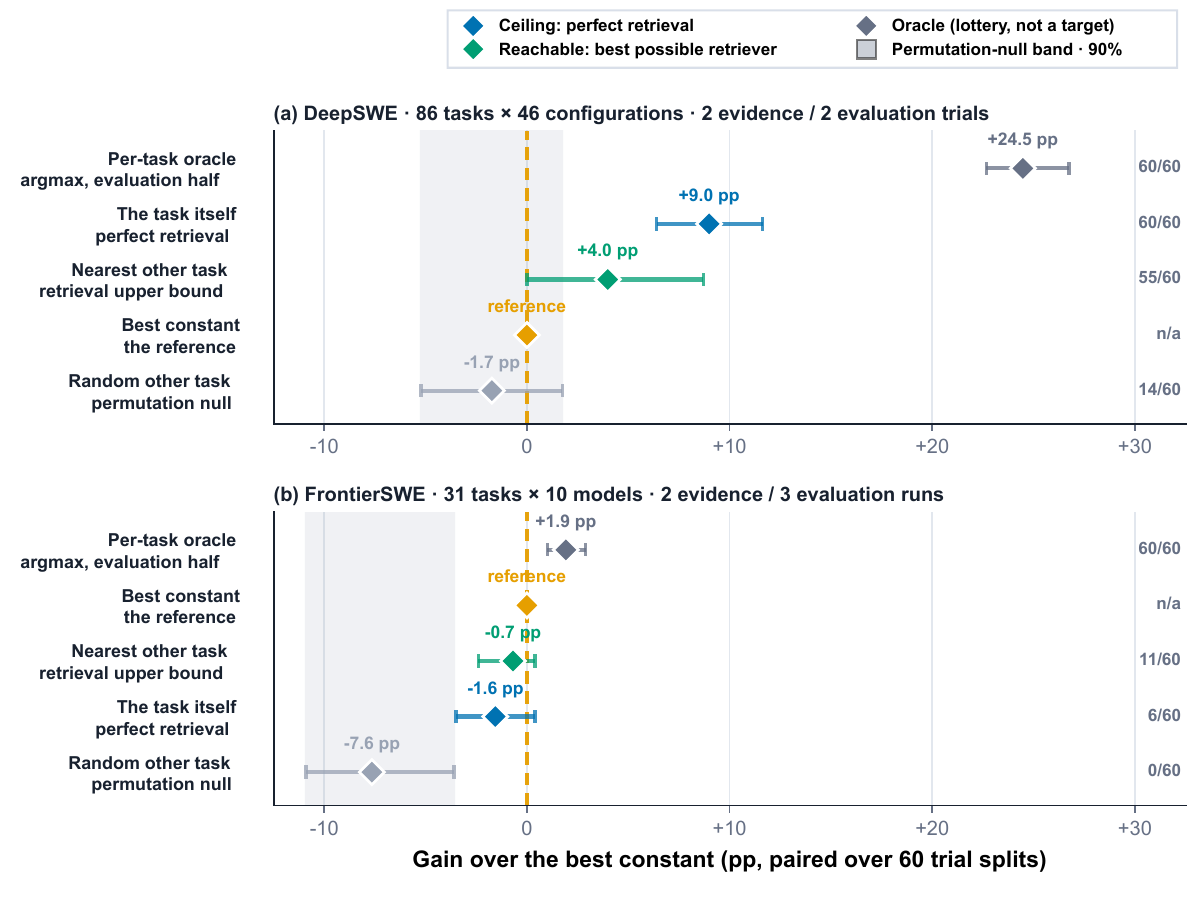}
\caption{Accuracy gains from different sources of routing evidence.}
\label{fig:ceiling}
\end{figure}

\end{document}